\pdfoutput=1

\documentclass[11pt]{article}

\usepackage[]{EMNLP2022}

\usepackage{times}
\usepackage{latexsym}

\usepackage{multirow, booktabs}
\usepackage{amsmath}
\usepackage{amssymb}
\usepackage{mathtools}
\usepackage{amsthm}
\usepackage{subcaption}

\usepackage[T1]{fontenc}

\usepackage[utf8]{inputenc}

\usepackage{microtype}

\usepackage{inconsolata}

\title{Synergy with Translation Artifacts for\\Training and Inference in Multilingual Tasks}

\author{Jaehoon Oh\thanks{\;\;Equal contribution} \\ Graduate School of DS, KAIST \\ \texttt{jhoon.oh@kaist.ac.kr} \\
        \And  
        Jongwoo Ko$^*$,\;Se-Young Yun \\ Graduate School of AI, KAIST \\ \texttt{\{jongwoo.ko, yunseyoung\}@kaist.ac.kr}\\
        }

\begin{document}
\maketitle

\begin{abstract}
Translation has played a crucial role in improving the performance on multilingual tasks: (1) to generate the target language data from the source language data for training and (2) to generate the source language data from the target language data for inference. However, prior works have not considered the use of both translations simultaneously. This paper shows that combining them can synergize the results on various multilingual sentence classification tasks. We empirically find that translation artifacts stylized by translators are the main factor of the performance gain. Based on this analysis, we adopt two training methods, SupCon and MixUp, considering translation artifacts. Furthermore, we propose a cross-lingual fine-tuning algorithm called MUSC, which uses SupCon and MixUp jointly and improves the performance. Our code is available at \url{https://github.com/jongwooko/MUSC}.
\end{abstract}

\section{Introduction}\label{sec:intro}

Large-scale pre-trained multilingual language models\,\citep{devlin-etal-2019-bert,conneau2019cross,huang-etal-2019-unicoder,conneau-etal-2020-unsupervised,luo-etal-2021-veco} have shown promising transferability in zero-shot cross-lingual transfer (ZSXLT), where pre-trained language models (PLMs) are fine-tuned using a labeled task-specific dataset from a \emph{rich-resource} source language (e.g., English or Spanish) and then evaluated on \emph{zero-resource} target languages.
Multilingual PLMs yield a universal representation space across different languages, thereby improving multilingual task performance\,\citep{pires-etal-2019-multilingual,chen-etal-2019-multi-source}.
Recent work has enhanced cross-lingual transferability by reducing the discrepancies between languages based on translation approaches during fine-tuning\,\citep{fang2021filter,zheng-etal-2021-consistency,yang2022enhancing}.
Our paper focuses on when translated datasets are available for cross-lingual transfer\,(XLT).

\citet{conneau-etal-2018-xnli} provided two translation-based XLT baselines: \texttt{translate-train} and \texttt{translate-test}.
The former fine-tunes a multilingual PLM (e.g., multilingual BERT) using the \emph{original} source language and \emph{machine-translated} target languages simultaneously and then evaluates it on the target languages.
Meanwhile, the latter fine-tunes a source language-based PLM (e.g., English BERT) using the \emph{original} source language and then evaluates it on the \emph{machine-translated} source language. Both baselines improve the performance compared to ZSXLT; however, they are sensitive to the translator, including translation artifacts, which are characteristics stylized by the translator\,\citep{conneau-etal-2018-xnli, artetxe-etal-2020-translation}. 

\citet{artetxe-etal-2020-translation} showed that matching the types of text (i.e., origin or translationese\footnote{Original text is directly written by humans. Translationese includes both human-translated and machine-translated texts.}) between training and inference is essential due to the presence of translation artifacts under \texttt{translate-test}.
Recently, \citet{yu-etal-2022-translate} proposed a training method that projects the original and translated texts into the same representation space under \texttt{translate-train}.
However, prior works have not considered the two baselines simultaneously.

In this paper, we combine \texttt{translate-train} and \texttt{translate-test} using a pre-trained multilingual BERT, to improve the performance.
Next, we identify that fine-tuning using the \emph{translated target} dataset is required to improve the performance on the \emph{translated source} dataset due to translation artifacts even if the languages for training and inference are different.
Finally, to consider translation artifacts during fine-tuning, we adopt two training methods, supervised contrastive learning (SupCon; \citealt{khosla2020supervised}) and MixUp\,\citep{zhang2018mixup} and propose MUSC, which combines them and improves the performance for multilingual sentence classification tasks.
\section{Scope of the Study}\label{sec:scope}

\begin{table}[t]
    \centering
    \caption{Notations of datasets.}\label{tab:notation}
    \begin{tabular}{cl}
    \toprule
    Notation & Description \\
    \midrule
    $\mathcal{S}_{\sf trn}$    & given source dataset for training \\
    $\mathcal{T}_{\sf trn}$    & given target dataset for training \\
    \multirow{2}{*}{$\mathcal{T}_{\sf trn}^{\sf MT}$} & machine-translated target dataset \\
    & from $\mathcal{S}_{\sf trn}$ for training \\
    \multirow{2}{*}{$\mathcal{T}_{\sf trn}^{\sf BT}$} & back-translated target dataset \\
    & from $\mathcal{T}_{\sf trn}$ for training \\
    \midrule
    $\mathcal{T}_{\sf tst}$    & given target dataset for inference \\
    \multirow{2}{*}{$\mathcal{S}_{\sf tst}^{\sf MT}$} & machine-translated source dataset \\
    & from $\mathcal{T}_{\sf tst}$ for inference \\ 
    \bottomrule
    \end{tabular}
\end{table}

\begin{table}[t]
    \centering
    \caption{Algorithm comparison.}\label{tab:alg_data}
    \resizebox{\linewidth}{!}{%
    \begin{tabular}{c|ccc}
    \toprule
    Algorithm & PLM & Training & Inference \\ \midrule
    ZSXLT & Multilingual & $\mathcal{S}_{\sf trn}$ & $\mathcal{T}_{\sf tst}$ \\
    \texttt{translate-train} & Multilingual & $\mathcal{S}_{\sf trn}$ \& $\mathcal{T}_{\sf trn}^{\sf MT}$ & $\mathcal{T}_{\sf tst}$ \\
    \texttt{translate-test} & English & $\mathcal{S}_{\sf trn}$ & $\mathcal{S}_{\sf tst}^{\sf MT}$ \\
    \texttt{translate-all} & Multilingual & $\mathcal{S}_{\sf trn}$ \& $\mathcal{T}_{\sf trn}^{\sf MT}$ & $\mathcal{T}_{\sf tst}$ \& $\mathcal{S}_{\sf tst}^{\sf MT}$ \\
    \bottomrule
    \end{tabular}}
\end{table}

In this study, four datasets are used: MARC and MLDoc for single sentence classification, and PAWSX and XNLI from XTREME\,\citep{pmlr-v119-hu20b} for sentence pair classification. The details of datasets are provided in Appendix \ref{appx:dataset}.
Each dataset consists of the source dataset for training $\mathcal{S}_{\sf trn}$ and the target dataset for inference $\mathcal{T}_{\sf tst}$, where $\mathcal{S}_{\sf trn}$ is original and $\mathcal{T}_{\sf tst}$ is original (for MARC and MLDoc) or human-translated (for PAWSX and XNLI).
For MARC and MLDoc, the original target dataset for training $\mathcal{T}_{\sf trn}$ is additionally given.
We use the given translated datasets $\mathcal{T}_{\sf trn}^{\sf MT}$ for PAWSX and XNLI. However, for MARC and MLDoc, the translated datasets are not given. Therefore, we use an m2m\_100\_418M translator\,\citep{fan2021beyond} from the open-source library \texttt{EasyNMT}\footnote{https://github.com/UKPLab/EasyNMT} to create the translated datasets.
$\mathcal{T}_{\sf trn}^{\sf MT}$is translated from $\mathcal{S}_{\sf trn}$ (i.e., $\mathcal{S}_{\sf trn} \rightarrow \mathcal{T}_{\sf trn}^{\sf MT}$), and $\mathcal{T}_{\sf trn}^{\sf BT}$ is back-translated from $\mathcal{T}_{\sf trn}$ (i.e., $\mathcal{T}_{\sf trn} \rightarrow \mathcal{S}_{\sf trn}^{\sf MT} \rightarrow \mathcal{T}_{\sf trn}^{\sf BT}$; \citealt{sennrich-etal-2016-improving}).
Similarly, for inference, $\mathcal{S}_{\sf tst}^{\sf MT}$ is translated from $\mathcal{T}_{\sf tst}$.
The notations used in this paper are listed in Table\,\ref{tab:notation}.

We use the pre-trained cased multilingual BERT\,\citep{devlin-etal-2019-bert} from HuggingFace Transformers\,\citep{wolf-etal-2020-transformers} and use accuracy as a metric. Detailed information for fine-tuning is provided in Appendix \ref{appx:implement}.

\section{Original and Translationese Ensemble}\label{sec:experiments}

\begin{table}[t]
\centering
\caption{Results according to the inference datasets (Acc. in \%). $\mathcal{S}_{\sf trn}$ and $\mathcal{T}_{\sf trn}^{\sf MT}$ are used for training. The number in the parenthesis of MLDoc is the number of training samples. `Ens.' indicates the ensemble of results on the two different test datasets in the inference. XNLI results are reported in Appendix \ref{appx:xnli}.}\label{tab:all_enc}
\addtolength{\tabcolsep}{-4.0pt}
\resizebox{\linewidth}{!}{%
\begin{tabular}{cc|ccccccccc|c}
    \toprule
    Dataset & Inference   & \textbf{EN}  & \textbf{ZH} & \textbf{FR} & \textbf{DE} & \textbf{RU} & \textbf{ES} & \textbf{IT} & \textbf{KO} & \textbf{JA} & \textbf{Avg.} \\ \midrule
    \multirow{3}{*}{MARC} & {$\mathcal{T}_{\sf tst}$} & 65.2 & 47.8 & 55.4 & 59.1 & - & 55.8 & - & - & 47.8 & 55.1 \\
    & {$\mathcal{S}_{\sf tst}^{\sf MT}$} & 65.2 & 44.9 & 54.4 & 59.8 & - & 55.4 & - & - & 44.9 & 54.5 \\
    & Ens. & 65.2 & 49.3 & 56.1 & 61.2 & - & 56.2 & - & - & 48.8 & \textbf{56.1} \\  \midrule
    \multirow{3}{*}{\shortstack{MLDoc\\(1000)}} & {$\mathcal{T}_{\sf tst}$} & 91.1 & 77.4 & 74.5 & 84.0 & 67.9 & 74.4 & 65.0 & - & 74.4 & 76.1 \\
    & {$\mathcal{S}_{\sf tst}^{\sf MT}$} & 91.1 & 77.6 & 79.0 & 88.1 & 61.3 & 76.4 & 72.3 & - & 67.3 & 76.6 \\
    & Ens. & 91.1 & 78.9 & 78.3 & 87.9 & 66.1 & 76.2 & 71.2 & - & 74.9 & \textbf{78.1} \\ \midrule
    \multirow{3}{*}{\shortstack{MLDoc\\(10000)}} & {$\mathcal{T}_{\sf tst}$} & 97.4 & 82.6 & 91.1 & 91.0 & 72.2 & 85.9 & 78.0 & - & 72.6 & 83.8\\
    & {$\mathcal{S}_{\sf tst}^{\sf MT}$} & 97.4 & 86.4 & 92.0 & 92.6 & 72.4 & 88.2 & 79.0 & - & 71.0 & 84.9 \\
    & Ens. & 97.4 & 87.7 & 92.2 & 92.6 & 72.1 & 88.0 & 80.6 & - & 75.9 & \textbf{85.8} \\ \midrule
    \multirow{3}{*}{PAWSX} & {$\mathcal{T}_{\sf tst}$} &  94.5 & 85.0 & 91.2 & 89.0 & - & 90.5 & - & 83.1 & 83.3 & 88.1 \\ 
    & {$\mathcal{S}_{\sf tst}^{\sf MT}$} & 94.5 & 84.5 & 91.7 & 90.6 & - & 91.3 & - & 83.1 & 80.9 & 88.1 \\ 
    & Ens. & 94.5 & 86.1 & 92.0 & 91.2 & - & 91.6 & - & 85.3 & 82.8 & \textbf{89.1} \\ \bottomrule
    \end{tabular}
} 
\end{table}


In this section, we demonstrate that the two baselines, \texttt{translate-train} and \texttt{translate-test}, are easily combined to improve performance, which we call it \texttt{translate-all}. Table\,\ref{tab:alg_data} describes the differences between algorithms.

Table~\ref{tab:all_enc} presents the results according to the inference dataset when the models are fine-tuned using $\mathcal{S}_{\sf trn}$ and $\mathcal{T}_{\sf trn}^{\sf MT}$. Inference on $\mathcal{T}_{\sf tst}$ is a general way to evaluate the models, i.e., \texttt{translate-train}.
In addition, we evaluate the models on $\mathcal{S}_{\sf tst}^{\sf MT}$ like \texttt{translate-test}. Furthermore, we ensemble the two results from different test datasets by averaging the predicted predictions, i.e., \texttt{translate-all}, because averaging the predictions over models or data points is widely used to improve predictive performance and uncertainty estimation of models\,\citep{gontijo-lopes2022no, kim2020learning}.

From Table~\ref{tab:all_enc}, it is shown that even if the multilingual PLMs are fine-tuned with $\mathcal{S}_{\sf trn}$ and $\mathcal{T}_{\sf trn}^{\sf MT}$, the performance on the translated source data $\mathcal{S}_{\sf tst}^{\sf MT}$ is competitive with that on the target data $\mathcal{T}_{\sf tst}$.
Furthermore, ensemble inference increases the performance on all datasets.
This can be interpreted as the effectiveness of the test time augmentation\,\citep{kim2020learning, ashukha2021mean}, because the results on the two \emph{test} datasets, $\mathcal{T}_{\sf tst}$ and $\mathcal{S}_{\sf tst}^{\sf MT}$ (augmented from $\mathcal{T}_{\sf tst}$), are combined.

\begin{figure}[t]
    \hspace*{\fill}
    \begin{subfigure}[b]{0.40\linewidth}
    \centering
    \includegraphics[width=\linewidth]{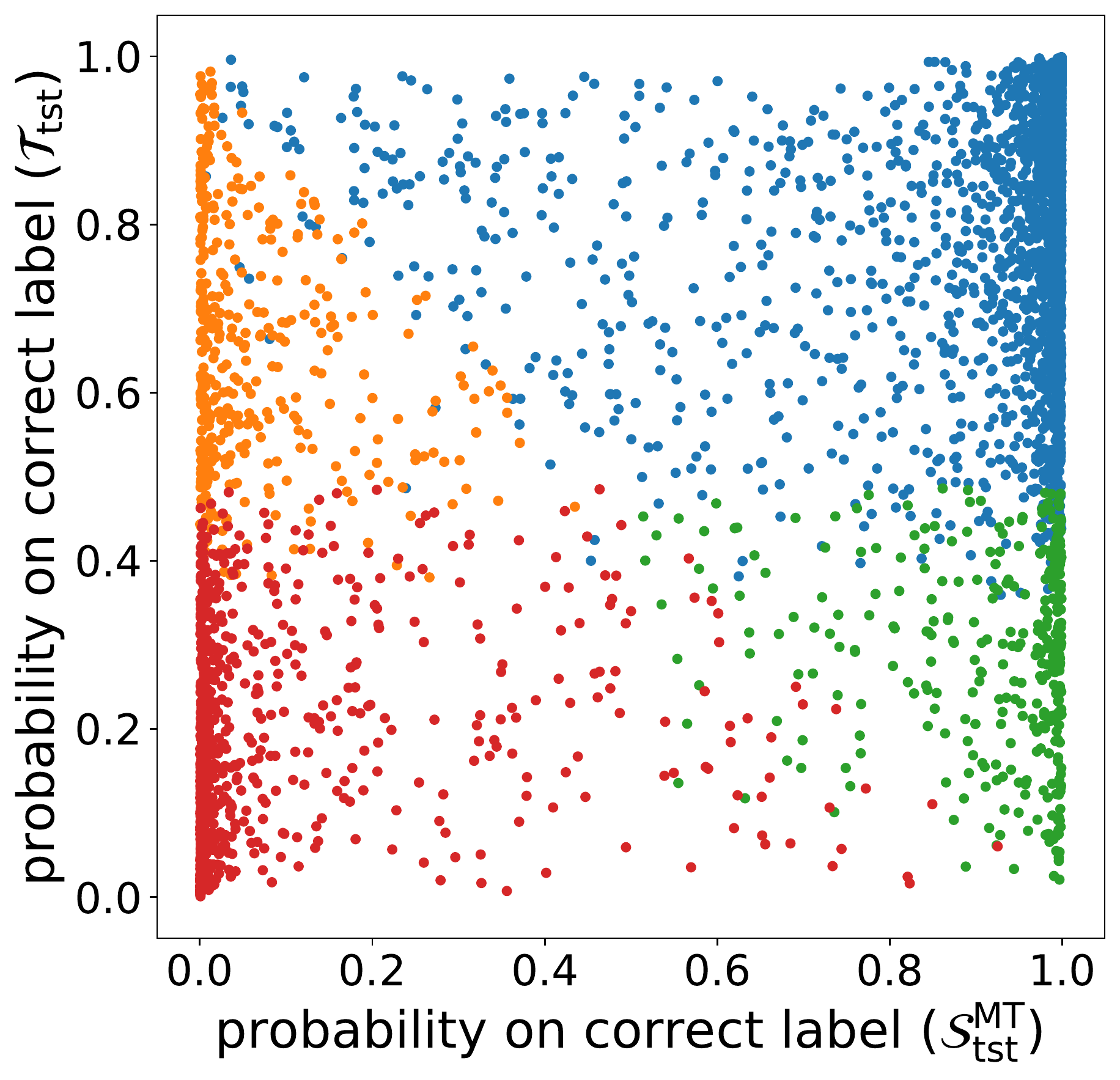}
    \caption{ZH on XNLI}
    \end{subfigure}
    \hfill
    \begin{subfigure}[b]{0.40\linewidth}
    \centering
    \includegraphics[width=\linewidth]{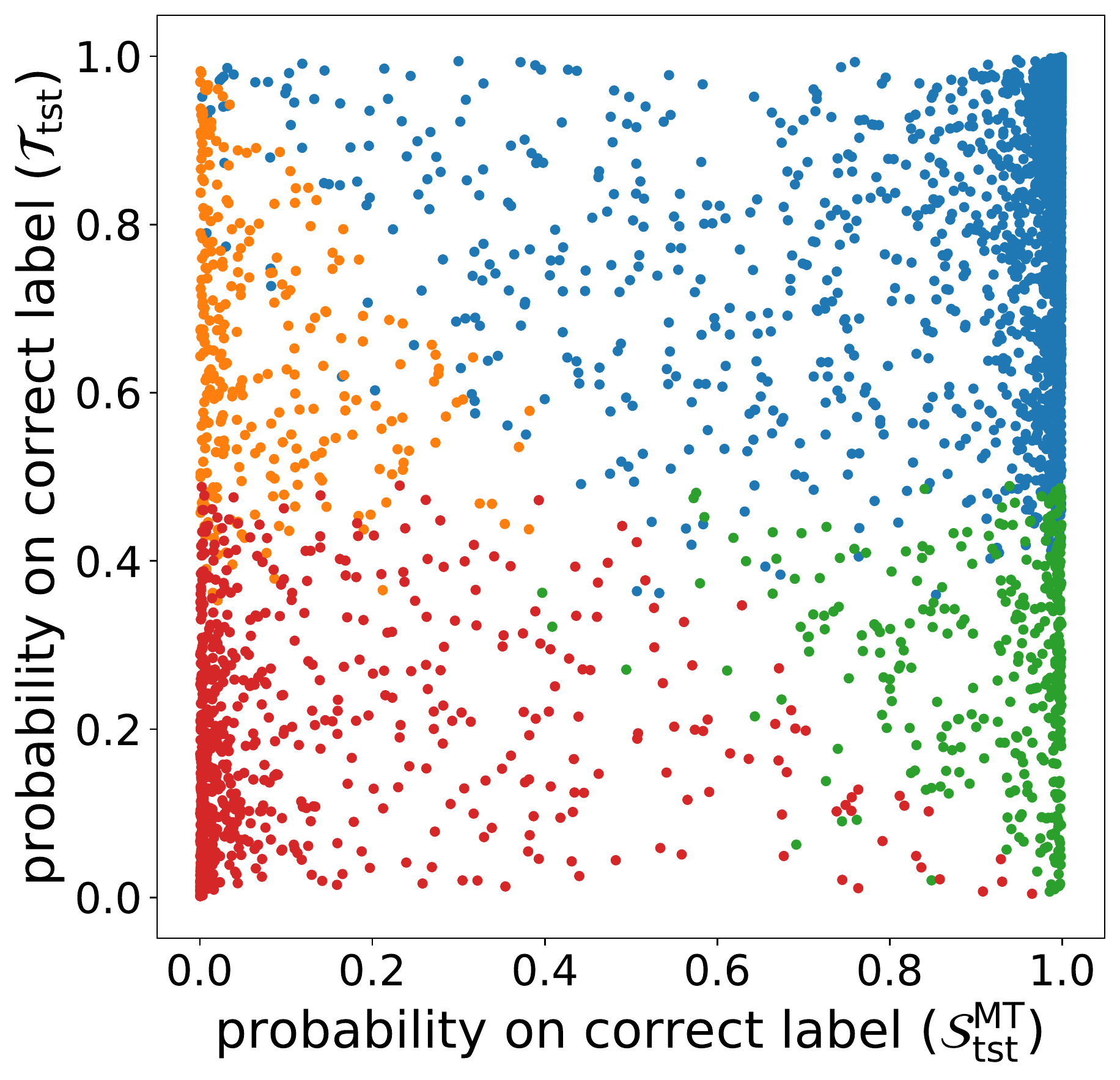}
    \caption{DE on XNLI}
    \end{subfigure}
    \hspace*{\fill}
    \caption{Predicted probability values on correct label when the models are evaluated on $\mathcal{T}_{\sf tst}$ and $\mathcal{S}_{\sf tst}^{\sf MT}$. The colors indicate right or wrong predictions: right on $\mathcal{T}_{\sf tst}$ and right on Ens. (blue), right on $\mathcal{T}_{\sf tst}$ and wrong on Ens. (orange), wrong on $\mathcal{T}_{\sf tst}$ and right on Ens. (green), and wrong on $\mathcal{T}_{\sf tst}$ and wrong on Ens. (red).}
    \label{fig:ens_figure}
\end{figure}

To explain the changes in inferences via test time augmentation, we describe the predicted probability values on the correct label when the models are evaluated on $\mathcal{T}_{\sf tst}$ and $\mathcal{S}_{\sf tst}^{\sf MT}$, as depicted in Figure~\ref{fig:ens_figure}. The green and orange dots represent the benefits and losses via the ensemble, respectively. The improved performance through the ensemble means that the number of green samples is greater than the number of orange samples in Figure~\ref{fig:ens_figure}.

To analyze where the performance gain comes from, we focus on the green samples. The green samples are concentrated around the right down corner, which implies that wrong predictions on $\mathcal{T}_{\sf tst}$ can be right predictions with high confidence on $\mathcal{S}_{\sf tst}^{\sf MT}$.
In fact, this phenomenon is the opposite of what we expected; the samples are expected to be concentrated around the $y=x$ line, because the semantic meaning between $\mathcal{T}_{\sf tst}$ and $\mathcal{S}_{\sf tst}^{\sf MT}$ is similar even though the languages are different.
This implies that semantic meaning is not the main factor explaining the performance gain of the ensemble.
\section{Translation Artifacts for Training}\label{sec:method1}

\begin{table}[t]
\centering
\caption{Results according to the matching between types of text for training and inference (Acc. in \%). $\mathcal{S}_{\sf trn}$ is also used for training.}\label{tab:all_bt}
\addtolength{\tabcolsep}{-4.0pt}
\resizebox{\linewidth}{!}{%
\begin{tabular}{ccc|cccccccc|c}
    \toprule
    Dataset & Training & Inference & \textbf{EN}  & \textbf{ZH} & \textbf{FR} & \textbf{DE} & \textbf{RU} & \textbf{ES} & \textbf{IT} & \textbf{JA} & \textbf{Avg.} \\ \midrule
    \multirow{4}{*}{MARC} & $\mathcal{T}_{\sf trn}$ & \multirow{2}{*}{$\mathcal{T}_{\sf tst}$} & 65.3 & 57.9 & 61.4 & 65.5 & - & 62.0 & - & 60.1 & \textbf{62.0} \\
    & $\mathcal{T}_{\sf trn}^{\sf BT}$ &  & 65.7 & 55.7 & 60.1 & 63.9 & - & 60.3 & - & 56.7 & 60.4 \\ \cmidrule{2-12}
     & $\mathcal{T}_{\sf trn}$ & \multirow{2}{*}{$\mathcal{S}_{\sf tst}^{\sf MT}$} & 65.3 & 48.2 & 57.1 & 61.8 & - & 57.7 & - & 47.1 & 56.2 \\
    & $\mathcal{T}_{\sf trn}^{\sf BT}$ &  & 65.7 & 49.2 & 57.7 & 62.4 & - & 57.2 & - & 48.5 & \textbf{56.8} \\ \midrule
    \multirow{4}{*}{\shortstack{MLDoc\\(1000)}} & $\mathcal{T}_{\sf trn}$ & \multirow{2}{*}{$\mathcal{T}_{\sf tst}$} & 93.7 & 91.9 & 93.6 & 95.6 & 87.2 & 95.5 & 86.8 & 89.3 & \textbf{91.7} \\
     & $\mathcal{T}_{\sf trn}^{\sf BT}$ &  & 93.4 & 90.6 & 93.5 & 95.1 & 87.1 & 92.7 & 86.4 & 86.4 & 90.6 \\ \cmidrule{2-12}
     & $\mathcal{T}_{\sf trn}$ & \multirow{2}{*}{$\mathcal{S}_{\sf tst}^{\sf MT}$} & 93.7 & 86.4 & 92.5 & 93.9 & 83.8 & 93.1 & 80.6 & 73.3 & 87.2 \\
    & $\mathcal{T}_{\sf trn}^{\sf BT}$ &  & 93.4 & 87.2 & 93.0 & 94.8 & 84.0 & 93.2 & 80.5 & 75.9 & \textbf{87.7} \\ \midrule
    \multirow{4}{*}{\shortstack{MLDoc\\(10000)}} & $\mathcal{T}_{\sf trn}$ & \multirow{2}{*}{$\mathcal{T}_{\sf tst}$} & 96.8 & 93.9 & 96.7 & 97.5 & 89.5 & 96.8 & 92.2 & 92.5 & \textbf{94.5} \\
    & $\mathcal{T}_{\sf trn}^{\sf BT}$ &  & 97.0 & 93.3 & 96.1 & 97.2 & 87.9 & 95.7 & 90.8 & 89.5 & 93.4 \\ \cmidrule{2-12}
     & $\mathcal{T}_{\sf trn}$ & \multirow{2}{*}{$\mathcal{S}_{\sf tst}^{\sf MT}$} & 96.8 & 88.9 & 94.9 & 96.4 & 84.3 & 94.0 & 83.5 & 75.7 & 89.3 \\
    & $\mathcal{T}_{\sf trn}^{\sf BT}$ &  & 97.0 & 87.6 & 94.6 & 95.4 & 84.2 & 93.8 & 85.7 & 77.2 & \textbf{89.4} \\ \bottomrule
    \end{tabular}
} 
\end{table}

%
To find the main factor of performance gain, we hypothesize that matching the types of text (i.e., original or translated) between training and inference is important \emph{even if the languages used for training and inference are different}, by expanding on \citet{artetxe-etal-2020-translation}.
For the analysis, we use MARC and MLDoc because they provide $\mathcal{T}_{\sf trn}$, which has no artifacts.

Table~\ref{tab:all_bt} describes the results according to the matching between texts for training and inference.
Well-matched texts are better than badly matched ones.
In particular, the results that $\mathcal{T}_{\sf trn}^{\sf BT}$--$\mathcal{S}_{\sf tst}^{\sf MT}$ is better than $\mathcal{T}_{\sf trn}$--$\mathcal{S}_{\sf tst}^{\sf MT}$ support our hypothesis.
This implies that biasing training and inference datasets using the same translator can lead to performance improvement, and that translation artifacts can change wrong predictions on $\mathcal{T}_{\sf tst}$ into right predictions on $\mathcal{S}_{\sf tst}^{\sf MT}$ when the models are trained using $\mathcal{T}_{\sf trn}^{\sf MT}$, as shown in Section \ref{sec:experiments}.
\begin{figure*}[t]
    \centering
    \includegraphics[width=1.0\linewidth]{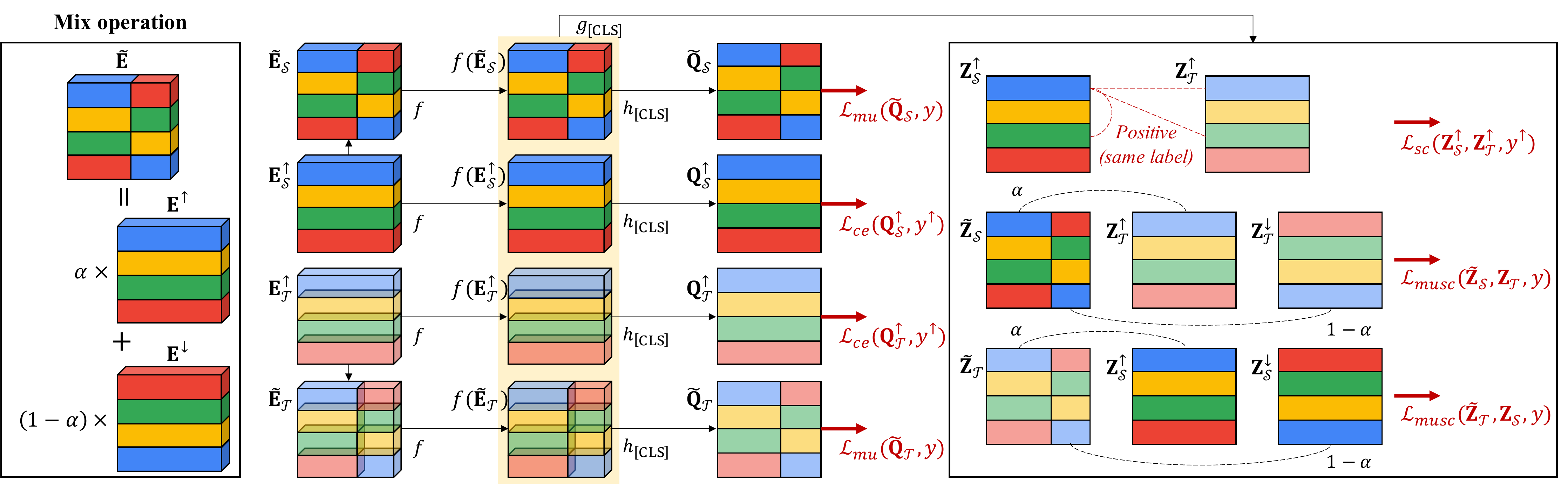}
    \caption{Overview of MUSC. $\text{\textbf{E}}_\mathcal{S}$ and $\text{\textbf{E}}_\mathcal{T}$ are the embeddings of the paired source and target languages, and each row indicates one sentence. Note that in the mix operation, addition and multiplication are operated elementwisely. $f$, $g$, and $h$ are the encoder, projector, and classifier, respectively. $g_{\text{[CLS]}} (f(\text{\textbf{E}}))$ and $h_{\text{[CLS]}} (f(\text{\textbf{E}}))$ means $g(f(\text{\textbf{E}})^{\text{[CLS]}})$ and $h(f(\text{\textbf{E}})^{\text{[CLS]}})$, respectively. $g(f(\text{\textbf{E}})^{\text{[CLS]}})$ is expressed as $\text{\textbf{Z}}$. In this figure, it is assumed that the batch size is four and that the blue- and green-colored samples have the same class.}
    \label{fig:concept}
\end{figure*}

\subsection{Proposed Method: MUSC}
We propose an XLT method called MUSC, by applying SupCon\,\citep{khosla2020supervised} and MixUp\,\citep{zhang2018mixup} jointly.
Namely, our method is contrastive learning with mixture sentences in \emph{supervised} settings.
Several works have attempted to employ the idea of mixtures on unsupervised contrastive learning \cite{kim2020mixco, shen2022unmix}; however, ours is the first to leverage the label information in a mixture.
In this section, the loss functions are formulated at batch level with a batch size of $N$, and $\uparrow$ and $\downarrow$ indicate the normal and reverse order, respectively, in a batch.
All methods are designed upon the \texttt{translate-all}.

\smallskip
\noindent\textbf{SupCon.} We adopt SupCon, which makes the samples in the same class closer\,\citep{gunel2021supervised}, to reduce the discrepancies between original and translated texts. Namely, SupCon helps models to learn both originality of $\mathcal{S}_{\sf trn}$ and artifacts of $\mathcal{T}_{\sf trn}^{\sf MT}$ comprehensively. The loss function of SupCon ($\mathcal{L}_{sc}$) with $I \equiv \left[ 1, \ldots, 2N \right]$ is as follows:
\begin{align*}
\resizebox{\linewidth}{!}{
    $\mathcal{L}_{sc} (\mathbf{Z}_{\mathcal{S}}, \mathbf{Z}_{\mathcal{T}}, \mathbf{y}) = \sum\limits_{i \in I} -\log \left\{ \frac{1}{\vert P(i) \vert} \sum\limits_{p \in P(i)} \frac{\exp (\mathbf{z}_i \cdot \mathbf{z}_{p} / \tau)}{\sum\limits_{\mathbf{z}_{a} \in \mathbf{Z}\setminus\{\mathbf{z}_{i}\}} \exp (\mathbf{z}_i \cdot \mathbf{z}_{a} / \tau)} \right\}$,
}
\end{align*}

\noindent where $\mathbf{Z} = \left[ \mathbf{Z}_{\mathcal{S}}; \mathbf{Z}_{\mathcal{T}} \right] \in \mathbb{R}^{2N \times d_p}$ is the projections of [CLS] token representations through an encoder $f$ and a projector $g$, i.e., $g(f(\mathbf{E})^{\text{[CLS]}})$, and $\mathbf{z}_i$ indicates the $i$-th row of $\mathbf{Z}$.
$\mathbf{Z}$ is concatenated along with the batch dimension and $d_p$ is the dimension of projections.
The positive set of the sample $i$, $P(i)$, is defined as $\{ j \vert y^{\prime}_{j} = y^{\prime}_{i}, j \in I \setminus \{i\} \}$, where $[y^{\prime}_{1}, \ldots, y^{\prime}_{N}] = [y^{\prime}_{N+1}, \ldots, y^{\prime}_{2N}] = \mathbf{y}$.

\smallskip
\noindent\textbf{MixUp.} We adopt MixUp to densify original and translated texts, respectively.
MixUp is performed on the word embeddings by following \citet{chen-etal-2020-mixtext}, because it is infeasible to directly apply MixUp to discrete word tokens. MixUp with $\alpha \in [0, 1]$ is as follows:
\begin{align*}
    \tilde{\mathbf{E}} = \text{\texttt{Mix}}_{\alpha}(\mathbf{E}^{\uparrow}, \mathbf{E}^{\downarrow}) = \alpha \mathbf{E}^{\uparrow} + (1 - \alpha) \mathbf
    {E}^{\downarrow}, 
\end{align*}
where $\mathbf{E}^{\uparrow} = \mathbf{X} \mathbf{W}_{e} \in \mathbb{R}^{N \times L \times d} $ is the output of the embedding layer for a given batch $\mathbf{X} \in \mathbb{R}^{N \times L \times \vert V \vert}$ with weight matrix $\mathbf{W}_{e} \in \mathbb{R}^{\vert V \vert \times d}$. $L$, $\vert V \vert$, and $d$ indicate maximum sequence length, vocab size, and dimension of word embeddings, respectively. $\mathbf{E}^{\downarrow}$ is reversed along with the batch dimension.
We apply MixUp between the same language to densify each type of text.
For convenience in implementation, we mix a normal batch $(\uparrow)$ and a reversed batch $(\downarrow)$, following \citet{shen2022unmix}. The mixing process is conducted elementwisely.
The loss function of MixUp ($\mathcal{L}_{mu}$) with cross-entropy ($\mathcal{L}_{ce}$) is as follows:
\begin{align*}
    \mathcal{L}_{mu}(\tilde{\mathbf{Q}}, \mathbf{y}) = \alpha \mathcal{L}_{ce} (\tilde{\mathbf{Q}}, \mathbf{y}^{\uparrow}) + (1 - \alpha) \mathcal{L}_{ce}(\tilde{\mathbf{Q}}, \mathbf{y}^{\downarrow}),
\end{align*}
where $\tilde{\mathbf{Q}} = h(f(\tilde{\mathbf{E}})^{\text{[CLS]}})$ is the logits of [CLS] token for the mixed embeddings, with an encoder $f$ and a classifier $h$. $\mathbf{y}$ is a set of labels in the same batch.

\smallskip
\noindent\textbf{MUSC.} We replace the original projected representations in $\mathcal{L}_{sc}$ with mixture ones, i.e., $\mathbf{Z}_{\mathcal{S}} \rightarrow \tilde{\mathbf{Z}}_{\mathcal{S}}$ or $\mathbf{Z}_{\mathcal{T}} \rightarrow \tilde{\mathbf{Z}}_{\mathcal{T}}$, to use MixUp and SupCon jointly. The loss functions of MUSC ($\mathcal{L}_{musc}$) are as follows:
\begin{align*}
    \resizebox{\linewidth}{!}{
    $\mathcal{L}_{musc} (\tilde{\mathbf{Z}}_{\mathcal{S}}, \mathbf{Z}_{\mathcal{T}}, \mathbf{y}) = \alpha \mathcal{L}_{sc} (\tilde{\mathbf{Z}}_{\mathcal{S}}, \mathbf{Z}_{\mathcal{T}}^{\uparrow}, \mathbf{y}^{\uparrow})  + (1 - \alpha) \mathcal{L}_{sc} (\tilde{\mathbf{Z}}_{\mathcal{S}}, \mathbf{Z}_{\mathcal{T}}^{\downarrow}, \mathbf{y}^{\downarrow})$,
    } \\
    \resizebox{\linewidth}{!}{
    $\mathcal{L}_{musc} (\tilde{\mathbf{Z}}_{\mathcal{T}}, \mathbf{Z}_{\mathcal{S}}, \mathbf{y}) = \alpha \mathcal{L}_{sc} (\tilde{\mathbf{Z}}_{\mathcal{T}}, \mathbf{Z}_{\mathcal{S}}^{\uparrow}, \mathbf{y}^{\uparrow})  + (1 - \alpha) \mathcal{L}_{sc} (\tilde{\mathbf{Z}}_{\mathcal{T}}, \mathbf{Z}_{\mathcal{S}}^{\downarrow}, \mathbf{y}^{\downarrow})$.
    }
\end{align*}

\noindent We calculate ${\mathcal{L}}_{musc}$ by decomposing it in two opposite orders, similar to ${\mathcal{L}}_{mu}$.
Finally, the total loss function ($\mathcal{L}$), descried in Figure~\ref{fig:concept}, is as follows:

\begin{table}[!ht]
\centering
\caption{Results according to the losses. $\mathcal{S}_{\sf trn}$ and $\mathcal{T}_{\sf trn}^{\sf MT}$ are used for training and $\mathcal{T}_{\sf tst}$ and $\mathcal{S}_{\sf tst}^{\sf MT}$ are used for ensemble inference, i.e., under \texttt{translate-all}. $-$ denotes baseline which only applies $\mathcal{L}_{ce}$. $\mathcal{L}_{ce}$ is basically added for all methods. XNLI results are reported in Appendix \ref{appx:xnli}.}\label{tab:all_musc}
\addtolength{\tabcolsep}{-4.5pt}
\resizebox{\linewidth}{!}{%
\begin{tabular}{cc|ccccccccc|c}
    \toprule
    Dataset & Method   & \textbf{EN} & \textbf{ZH} & \textbf{FR} & \textbf{DE} & \textbf{RU} & \textbf{ES} & \textbf{IT} & \textbf{KO} & \textbf{JA} & \textbf{Avg.} \\ \midrule
    \multirow{5}{*}{MARC} & $-$ & 65.2 & 49.3 & 56.1 & 61.2 & - & 56.2 & - & - & 48.8 & 56.1 \\
    & $\mathcal{L}_{sc}$ & 64.9 & 49.1 & 56.1 & 61.4 & - & 55.7 & - & - & 49.3 & 56.1 \\
    & $\mathcal{L}_{mu}$ & 64.5 & 49.4 & 55.5 & 61.5 & - & 55.9 & - & - & 48.7 & 55.9 \\
    & $\mathcal{L}_{sc}+\mathcal{L}_{mu}$ & 65.1 & 49.5 & 56.1 & 61.5 & - & 56.1 & - & - & 49.9 & \textbf{56.4} \\
    & $\mathcal{L}$ & 65.5 & 49.4 & 56.4 & 61.6 & - & 56.0 & - & - & 48.5 & 56.2 \\   \midrule
    \multirow{5}{*}{\shortstack{MLDoc\\(1000)}} & $-$ & 91.1 & 78.9 & 78.3 & 87.9 & 66.1 & 76.2 & 71.2 & - & 74.9 & 78.1 \\
    & $\mathcal{L}_{sc}$ & 95.0 & 86.0 & 85.0 & 91.4 & 67.3 & 84.2 & 75.0 & - & 72.3 & 82.0 \\
    & $\mathcal{L}_{mu}$ & 94.0 & 83.7 & 84.2 & 90.5 & 73.4 & 82.4 & 75.5 & - & 71.2 & 81.9 \\
    & $\mathcal{L}_{sc}+\mathcal{L}_{mu}$ & 91.7 & 85.0 & 88.6 & 90.4 & 71.0 & 82.9 & 76.7 & - & 75.3 & 82.7 \\
    & $\mathcal{L}$ & 94.8 & 86.7 & 86.2 & 90.2 & 73.3 & 80.8 & 74.8 & - & 77.5 & \textbf{83.1} \\ \midrule
    \multirow{5}{*}{\shortstack{MLDoc\\(2000)}} & $-$ & 95.7 & 87.3 & 85.9 & 91.3 & 80.5 & 81.4 & 76.7 & - & 78.2 & 84.6 \\
    & $\mathcal{L}_{sc}$ & 95.8 & 89.0 & 90.3 & 92.2 & 80.9 & 83.4 & 79.4 & - & 77.7 & 86.1 \\
    & $\mathcal{L}_{mu}$ & 95.9 & 88.8 & 92.3 & 92.3 & 81.1 & 85.9 & 78.5 & - & 75.7 & 86.3 \\
    & $\mathcal{L}_{sc}+\mathcal{L}_{mu}$ & 95.3 & 88.4 & 92.0 & 93.1 & 80.4 & 85.8 & 79.1 & - & 77.2 & 86.4 \\ 
    & $\mathcal{L}$ & 94.8 & 88.7 & 89.8 & 92.7 & 82.3 & 86.8 & 80.2 & - & 78.4 & \textbf{86.7} \\ \midrule
    \multirow{5}{*}{\shortstack{MLDoc\\(5000)}} & $-$ & 96.8 & 89.7 & 92.2 & 92.7 & 73.6 & 82.6 & 78.8 & - & 77.2 & 85.4 \\
    & $\mathcal{L}_{sc}$ & 96.7 & 89.0 & 93.0 & 93.7 & 71.5 & 88.2 & 81.1 & - & 77.6 & 86.3 \\
    & $\mathcal{L}_{mu}$ & 96.9 & 88.9 & 91.3 & 93.7 & 72.0 & 86.5 & 81.0 & - & 76.3 & 85.8 \\
    & $\mathcal{L}_{sc}+\mathcal{L}_{mu}$ & 96.6 & 88.1 & 92.2 & 92.2 & 78.7 & 85.4 & 80.2 & - & 76.3 & 86.2 \\
    & $\mathcal{L}$ & 97.0 & 88.6 & 92.9 & 95.0 & 77.7 & 89.1 & 81.1 & - & 72.9 & \textbf{86.8} \\ \midrule
    \multirow{5}{*}{\shortstack{MLDoc\\(10000)}} & $-$ & 97.4 & 87.7 & 92.2 & 92.6 & 72.1 & 88.0 & 80.6 & - & 75.9 & 85.8 \\
    & $\mathcal{L}_{sc}$ & 97.3 & 90.6 & 92.0 & 93.0 & 71.1 & 88.5 & 80.9 & - & 78.8 & 86.5 \\
    & $\mathcal{L}_{mu}$ & 97.4 & 88.7 & 92.5 & 94.8 & 71.6 & 89.9 & 79.4 & - & 72.7 & 85.9 \\
    & $\mathcal{L}_{sc}+\mathcal{L}_{mu}$ & 97.4 & 89.4 & 93.6 & 93.8 & 72.1 & 91.2 & 79.7 & - & 76.3 & 86.7 \\
    & $\mathcal{L}$ & 97.4 & 89.8 & 94.1 & 94.6 & 71.7 & 88.9 & 80.3 & - & 78.2 & \textbf{86.9} \\ \midrule
    \multirow{5}{*}{PAWSX} & $-$ &  94.5 & 86.1 & 92.0 & 91.2 & - & 91.6 & - & 85.3 & 82.8 & 89.1 \\ 
    & $\mathcal{L}_{sc}$ & 94.5 & 87.3 & 92.6 & 91.3 & - & 92.2 & - & 85.6 & 84.0 & 89.6 \\ 
    & $\mathcal{L}_{mu}$ & 94.5 & 86.3 & 92.3 & 92.2 & - & 92.4 & - & 85.3 & 84.8 & 89.7 \\ 
    & $\mathcal{L}_{sc}+\mathcal{L}_{mu}$ & 95.1 & 87.0 & 92.2 & 92.2 & - & 91.8 & - & 85.5 & 84.8 & \textbf{89.8} \\
    & $\mathcal{L}$ & 94.9 & 87.0 & 92.4 & 92.0 & - & 91.9 & - & 86.0 & 84.7 & \textbf{89.8} \\
    \bottomrule
    \end{tabular}
}
\end{table}

\begin{scriptsize}
\begin{align*}
    &\mathcal{L} = (1 - \lambda) \left [ \sum\limits_{i \in \left\{ \mathcal{S}, \mathcal{T} \right\}} \mathcal{L}_{c\!e}(\mathbf{Q}_{i}, \mathbf{y}) + \sum\limits_{i \in \left\{ \mathcal{S}, \mathcal{T} \right\}} \mathcal{L}_{m\!u}(\tilde{\mathbf{Q}}_{i}, \mathbf{y}) \right ] \\
    &\!\! + \lambda \left [ \mathcal{L}_{s\!c} (\mathbf{Z}_{\mathcal{S}}, \mathbf{Z}_{\mathcal{T}}, \mathbf{y}) + \mathcal{L}_{m\!u\!s\!c} (\tilde{\mathbf{Z}}_{\mathcal{S}}, \mathbf{Z}_{\mathcal{T}}, \mathbf{y}) + \mathcal{L}_{m\!u\!s\!c} (\tilde{\mathbf{Z}}_{\mathcal{T}}, \mathbf{Z}_{\mathcal{S}}, \mathbf{y}) \right ].
\end{align*}
\end{scriptsize}



Table \ref{tab:all_musc} describes the ablation study according to the applied loss functions. $-$ denotes baseline which only applies $\mathcal{L}_{ce}$. Other methods include $\mathcal{L}_{ce}$ and additionally apply the corresponding loss, respectively. It is shown that SupCon ($\mathcal{L}_{sc}$) and MixUp ($\mathcal{L}_{mu}$) improve performance on most datasets even when they are used separately. The effectiveness of these losses is powerful when dataset size is small. Moreover, our total loss ($\mathcal{L}$), which includes learning a model using SupCon and MixUp jointly ($\mathcal{L}_{musc}$), outperforms both SupCon and MixUp on all datasets. In addition, our total loss ($\mathcal{L}$) brings more performance gains than the simple conjunction of SupCon and MixUp ($\mathcal{L}_{sc}+\mathcal{L}_{mu}$) for all datasets except for MARC dataset. These results demonstrate that our proposed MUSC effectively collaborates the SupCon and MixUp. The optimized hyperparameters are reported in Appendix \ref{appx:implement}.

\section{Conclusion}\label{sec:conclusion}
In this paper, we showed that \texttt{translate-train} and \texttt{translate-test} are easily synergized from the test time augmentation perspective and found that the improved performance is based on translation artifacts.
Based on our analysis, we propose MUSC, which is supervised contrastive learning with mixture sentences, to enhance the generalizability on translation artifacts.
Our work highlighted the role of translation artifacts for XLT.

\section*{Limitations}
Our work addressed the role of translation artifacts for cross-lingual transfer. Limitation of our work is that we experimented for sentence classification tasks using multilingual BERT, because it is almost impossible to get token-level ground truths using translator.

\section*{Ethics Statement}
Our work does not violate the ethical issues. Furthermore, we showed that a new baseline, \texttt{translate-all}, could achieve higher performance, and proposed MUSC designed upon the \texttt{translate-all} approach. We believe that various algorithms can be developed based on the \texttt{translate-all} for multilingual tasks.

\section*{Acknowledgements}
This work was supported by Institute of Information \& communications Technology Planning \& Evaluation (IITP) grant funded by the Korea government(MSIT) (No.2019-0-00075, Artificial Intelligence Graduate School Program(KAIST), 10\%) and Institute of Information \& communications Technology Planning \& Evaluation(IITP) grant funded by the Korea government(MSIT) (No.2022-0-00641, XVoice: Multi-Modal Voice Meta Learning, 90\%)

\nocite{*} 
\bibliography{anthology,custom}
\bibliographystyle{acl_natbib}

\clearpage
\appendix

\section{Dataset Description}\label{appx:dataset}

\textbf{MARC}\,\citep{keung-etal-2020-multilingual} is Amazon review classification dataset. \textbf{MLDoc}\,\citep{schwenk2018corpus} is news article classification dataset. \textbf{PAWSX}\,\citep{yang-etal-2019-paws} is paraphrase identification dataset\footnote{https://console.cloud.google.com/storage/browser/\\xtreme\_translations/PAWSX}. \textbf{XNLI}\,\citep{conneau-etal-2018-xnli} is natural language inference dataset\footnote{https://console.cloud.google.com/storage/browser/\\xtreme\_translations/XNLI}.

\begin{table}[h]
    \centering
    \caption{Dataset description}\label{tab:dataset}
    \resizebox{\linewidth}{!}{%
    \begin{tabular}{c|ccccc}
    \toprule
    Dataset & \# of languages & \# of classes & \# of train & \# of val & \# of test \\ \midrule
    MARC & 6 & 5 & 200,000 & 5,000 & 5,000 \\
    MLDoc & 8 & 4 & 1,000-10,000 & 1,000 & 4,000 \\
    PAWSX & 7 & 2 & 49,401 & 2,000 & 2,000 \\
    XNLI & 15 & 3 & 392,702 & 2,490 & 5,010 \\
    \bottomrule
    \end{tabular}}
\end{table}

\section{Implementation Detail}\label{appx:implement}
Learning rate and $\lambda$ are searched by grid from [1e-5, 3e-5, 5e-5] and from [0.1, 0.5, 0.9], respectively. Fine-tuning epochs are 4, 10, 4, and 2 on MARC, MLDoc, PAWSX, and XNLI, respectively. The batch size is 32 for all datasets. The evaluation is executed every 300 batches on all languages. Table~\ref{tab:hyperparameters} describes the optimized hyperparameters.

\begin{table}[h]
    \centering
    \footnotesize
    \caption{Optimized hyperparameters}\label{tab:hyperparameters}
    \begin{tabular}{cc|cccc}
    \toprule
    & & MARC & MLDoc & PAWSX & XNLI \\ \midrule
    \multirow{2}{*}{$\mathcal{L}_{sc}$} & lr & 3e-5 & 3e-5 & 1e-5 & 3e-5 \\
     & $\lambda$ & 0.5 & 0.5/0.9 & 0.9 & 0.1 \\
    \midrule
    $\mathcal{L}_{mu}$ & lr & 3e-5 & 3e-5 & 1e-5 & 1e-5 \\
    \midrule
    \multirow{2}{*}{$\mathcal{L}$} & lr & 3e-5 & 3e-5 & 1e-5 & 1e-5 \\
     & $\lambda$ & 0.9 & 0.5/0.9 & 0.9 & 0.9 \\
    \bottomrule
    \end{tabular}
\end{table}

\section{XNLI results}\label{appx:xnli}

\begin{table}[h!]
\centering
\caption{XNLI results according to the inference datasets.}\label{tab:xnli_enc}
\resizebox{\linewidth}{!}{%
\begin{tabular}{c|ccccccccccccccc|c}
    \toprule
    Inference   & \textbf{EN} & \textbf{AR} & \textbf{BG} & \textbf{ZH} & \textbf{FR} & \textbf{DE} & \textbf{EL} & \textbf{HI} & \textbf{RU} & \textbf{ES} & \textbf{SW} & \textbf{TH} & \textbf{TR} & \textbf{UR} & \textbf{VI} & \textbf{Avg.} \\ \midrule
    {$\mathcal{T}_{\sf tst}$} & 82.2 & 73.1 & 77.8 & 77.6 & 78.2 & 77.3 & 75.0 & 71.0 & 76.0 & 79.3 & 67.6 & 67.8 & 73.3 & 67.2 & 76.8 & 74.7 \\ 
    {$\mathcal{S}_{\sf tst}^{\sf MT}$} & 82.2 & 74.9 & 78.4 & 75.2 & 77.9 & 78.4 & 77.5 & 71.5 & 75.3 & 78.6 & 69.5 & 71.3 & 75.7 & 67.3 & 75.0 & 75.3 \\
    Ens. & 82.2 & 76.7 & 79.6 & 77.3 & 79.1 & 79.8 & 78.8 & 73.1 & 77.3 & 79.9 & 71.1 & 73.0 & 77.6 & 68.8 & 77.1 & \textbf{76.8} \\ \bottomrule
    \end{tabular}
}
\end{table}

\begin{table}[h!]
\centering
\caption{XNLI results according to the training methods.}\label{tab:xnli_scmu}
\resizebox{\linewidth}{!}{
\begin{tabular}{c|ccccccccccccccc|c}
\toprule
Method & \textbf{EN} & \textbf{AR} & \textbf{BG} & \textbf{ZH} & \textbf{FR} & \textbf{DE} & \textbf{EL} & \textbf{HI} & \textbf{RU} & \textbf{ES} & \textbf{SW} & \textbf{TH} & \textbf{TR} & \textbf{UR} & \textbf{VI} & \textbf{Avg.} \\ \midrule
$-$ & 82.2 & 76.7 & 79.6 & 77.3 & 79.1 & 79.8 & 78.8 & 73.1 & 77.3 & 79.9 & 71.1 & 73.0 & 77.6 & 68.8 & 77.1 & 76.8 \\ 
$\mathcal{L}_{sc}$ & 82.7 & 77.2 & 79.8 & 77.7 & 80.0 & 80.4 & 79.7 & 73.7 & 76.6 & 80.7 & 72.7 & 73.5 & 77.7 & 69.8 & 77.0 & 77.3 \\ 
$\mathcal{L}_{mu}$ & 82.9 & 77.3 & 80.0 & 78.2 & 80.4 & 80.0 & 79.3 & 73.4 & 77.8 & 80.6 & 71.7 & 73.9 & 77.8 & 69.4 & 78.0 & 77.4 \\ 
$\mathcal{L}_{sc}+\mathcal{L}_{mu}$ & 83.5 & 77.6 & 79.8 & 78.4 & 80.6 & 80.5 & 80.0 & 73.5 & 78.1 & 80.9 & 72.5 & 73.7 & 77.8 & 69.9 & 77.6 & 77.6 \\
$\mathcal{L}$ & 83.6 & 77.9 & 80.2 & 78.1 & 80.8 & 80.3 & 79.7 & 73.6 & 78.0 & 81.0 & 72.2 & 73.6 & 77.8 & 70.4 & 77.8 & \textbf{77.7} \\
 \bottomrule
\end{tabular}
}
\end{table}

\section{Additional Related Works}\label{appx:related}
\textbf{Cross-lingual Transfer.} As the recent advances in NLP demonstrate the effectiveness of pre-trained language models (PLMs) like BERT\,\citep{devlin-etal-2019-bert} and RoBERTa\,\citep{liu2019roberta}, the performances of XLT rapidly improve by extending the monolingual PLMs to the multilingual settings\,\citep{conneau2019cross, conneau-etal-2020-unsupervised}.
While these multilingual PLMs show state-of-the-art performances in ZSXLT, one promising approach for improving the cross-lingual transferability is instance-based transfer by translation such as \texttt{translate-train} and \texttt{translate-test}\,\citep{conneau-etal-2018-xnli}. Due to the effectiveness and acceptability of translation, most recent works \citep{fang2021filter, zheng-etal-2021-consistency, yang2022enhancing} focus on better utilization of translation.

\smallskip
\noindent\textbf{Test-time augmentation.} Data augmentation, which expands a dataset by adding transformed copies of each example, is a common practice in supervised learning. While the data augmentation is also widely used in XLT\,\cite{zheng-etal-2021-consistency} during training models, it can also be used at the test time to obtain greater robustness\,\cite{prakash2018deflecting}, improved accuracy\,\cite{matsunaga2017image}, and estimates of uncertainty\,\cite{smith2018understanding}. Test time augmentation (TTA) combines predictions from a multi-viewed version of a single input to get a ``smoothed'' prediction. We also point out that using translation with XLT can be viewed as TTA, which can get performance gain from a different view of original and translation sentences. 
In this direction of the necessity of study for TTA\,\cite{kim2020learning}, we propose better utilization of translation artifacts in XLT.

\smallskip
\noindent\textbf{Translation artifacts.} ``Translationese'' can be referred to as characteristics in a translated text that differentiate it from the original text in the same language. While the effect of translationese has been widely studied in translation tasks \citep{graham2019translationese, freitag-etal-2020-bleu}, the efficacy of translationese in XLT is under-explored. \citet{artetxe-etal-2020-translation} and \citet{kaneko-bollegala-2021-debiasing} investigate the effect of translationese in \texttt{translate-test} and ZSXLT settings, however, these are apart from general training approach of XLT. Recently, \citet{yu-etal-2022-translate} firstly attempt to study \texttt{translate-train}, which focuses on single QA task.

\end{document}